\documentclass[letterpaper, 10 pt, conference]{ieeeconf}  

\IEEEoverridecommandlockouts                              

\pdfoutput=1

\usepackage{times,epsfig,graphicx,amsmath,amssymb,bm,amsfonts, comment, yhmath, booktabs, subfigure, soul, color, xcolor,threeparttable}
\soulregister\cite7
\soulregister\ref7
\soulregister\eqref7

\usepackage[pagebackref=true,breaklinks=true,colorlinks,bookmarks=false]{hyperref}

\title{\LARGE \bf
Phase-SLAM: Phase Based Simultaneous Localization \\ and Mapping for Mobile Structured Light Illumination Systems
}

\author{Xi Zheng$^{1}$, Rui Ma$^{1}$, Rui Gao$^{1}$, Qi Hao$^{* 1,2}$
	\thanks{This work is partially supported by the National Natural Science Foundation of China (No:  61773197);  and the Shenzhen Nanshan District Science and Technology Innovation Bureau (No: LHTD20170007); and the Intel ICRI-IACV Research Fund (No: CG\#52514373).}
	\thanks{$^{1}$ Xi Zheng, Rui Ma, Rui Gao, and Qi Hao are with the Department of Computer Science and Engineering, Southern University of Science and Technology, Shenzhen, China. 
	{\tt\small zhengx3@mail.sustech.edu.cn},
	{\tt\small mar@sustech.edu.cn}, 
	{\tt\small 12032493@mail.sustech.edu.cn}}%
    \thanks{$^{2}$ Qi Hao is with the Research Institute of Trustworthy Autonomous Systems, Southern University of Science and Technology, Shenzhen, China.}
	\thanks{$^{*}$ Corresponding author: Qi Hao ({\tt\small haoq@sustech.edu.cn})}%
}

\begin{document}
	
	\maketitle
	\thispagestyle{empty}
	\pagestyle{empty}	
	
\begin{abstract}
        
Structured Light Illumination (SLI) systems have been used for reliable indoor dense 3D scanning via phase triangulation. 
However, mobile SLI systems for 360 degree 3D reconstruction demand 3D point cloud registration, involving high computational complexity. 
In this paper, we propose a phase based Simultaneous Localization and Mapping (Phase-SLAM) framework for fast and accurate SLI sensor pose estimation and 3D object reconstruction. 
The novelty of this work is threefold:
(1) developing a reprojection model from 3D points to 2D phase data towards phase  registration with low computational complexity;
(2) developing a local optimizer to achieve SLI sensor pose estimation (odometry) using the derived Jacobian matrix for the 6 DoF variables; 
(3) developing a compressive phase comparison method to achieve high-efficiency loop closure detection. 
The whole Phase-SLAM pipeline is then exploited using existing global pose graph optimization techniques. 
We build datasets from both the unreal simulation platform and a robotic arm based SLI system in real-world to verify the proposed approach. 
The experiment results demonstrate that the proposed Phase-SLAM outperforms other state-of-the-art methods in terms of the efficiency and accuracy of pose estimation and 3D reconstruction. 
The open-source code is available at https://github.com/ZHENGXi-git/Phase-SLAM.
\end{abstract}
	
\section{INTRODUCTION}

The SLI technology has been widely used for high-precision 3D scanning for many industrial applications with the camera-projector pair. 
There are usually two approaches for SLI systems to achieve 360 degree 3D reconstruction: controlled motion based and free motion based \cite{salvi2004pattern}.
The former uses a servo motor to rotate the object along a pre-defined trajectory for multiple view scanning;
the latter estimates sensor motions through local and global point cloud registration, such as Iterative Closest Point (ICP) and associated variants \cite{ICP1992, low2004linear}. 
The free-motion approach is advantageous in its flexibility but incurs high computational complexity and demands a high storage capacity.
\begin{figure}
	\centering
	\vspace{-0.0cm}
	\setlength{\abovecaptionskip}{-3pt}
	\includegraphics[width=0.5\textwidth]{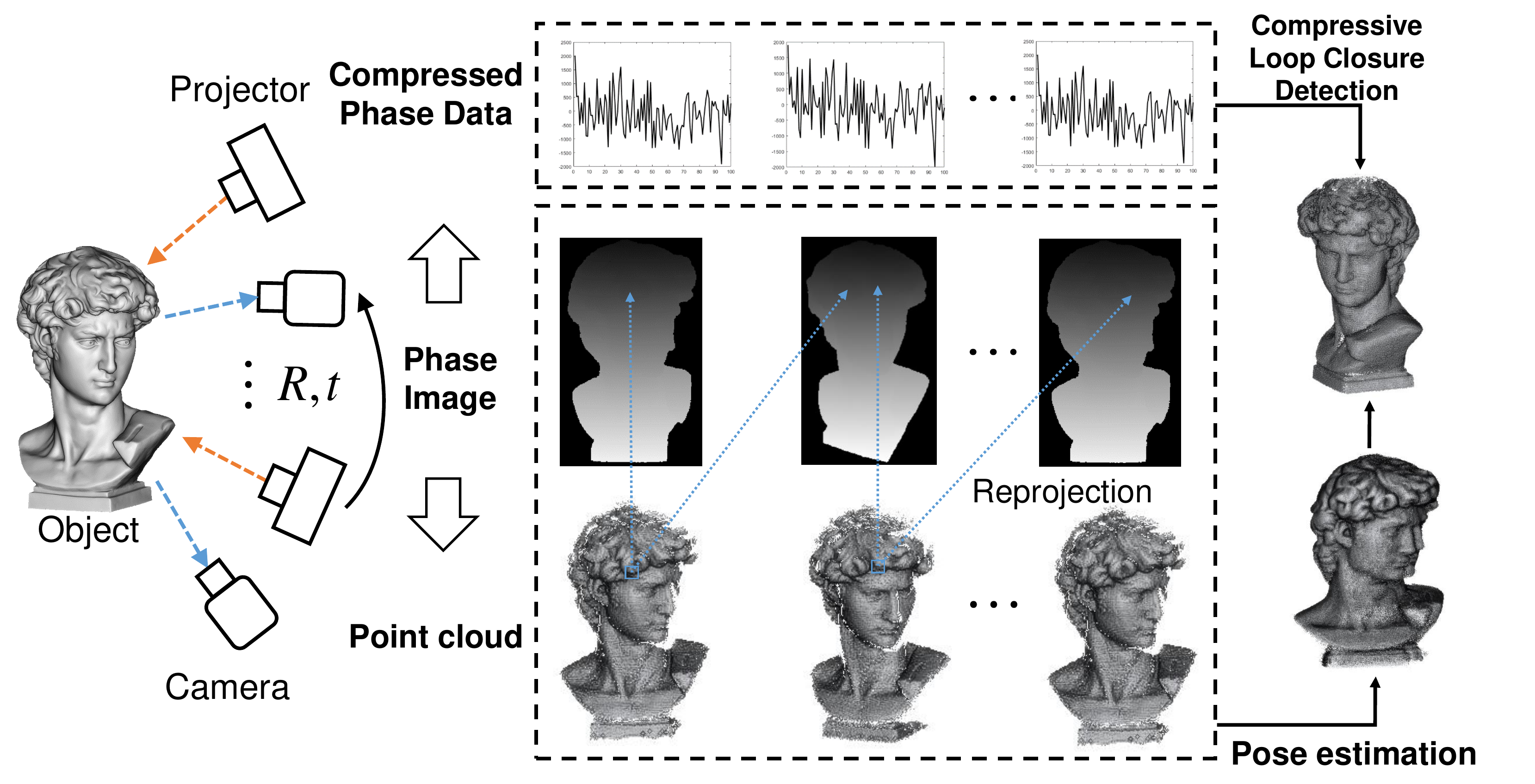}
	\caption{A diagram of the proposed Phase-SLAM framework based on the camera-projector pair, 
		which utilizes a 3D-to-2D reprojection model to predict the phase data for an assumed sensor pose, 
		a local optimizer to achieve pose estimation, and a compressive method to enable fast loop closure detection.}
	\label{fig: screen}
\end{figure}

Meanwhile, as the 2D phase data produced by SLI systems contain 3D information \cite{wang2012robust}, it is appealing to utilize the phase to achieve high-efficiency pose estimation and loop closure detection. 
However, to develop a fully functional Phase-SLAM system has to cope with the following technological challenges:
(1) how to build the intrinsic relationship between the phase and the transformation of SLI;
(2) how to develop a local optimization procedure for estimating 6 DOF motions of the SLI sensor (odometry); 
(3) how to achieve sparse representation and fast matching of phase data for loop closure detection. 
Our previous work \cite{zheng2021phase} proposes a geometric reference plane to model the relationship between phases and motions of 6 DoF separately, which is complicated and inconvenient.
Besides, if the loop closure detection is based on whole phase images, the memory footprint will also grow quickly as the scanning view increases.
This paper presents an upgraded Phase-SLAM framework, which utilizes a 3D point to 2D phase reprojection method to build the model, a gradient based local optimizer to achieve odometry functionality and a compression method to enable efficient loop detection (Fig. \ref{fig: screen}).
The main contributions of this work include,
\begin{enumerate}
	\item proposing a reprojection model 
	from 3D point to 2D phase data, which can be used to get phase estimations and measurements;
	\item constructing a local pose optimizer with the reprojection model and the analytical expression of Jacobian matrix is derived for pose estimation;
	\item developing a complete pipeline of Phase-SLAM framework with a compressive loop closure detection scheme and the pose graph optimization;
	\item building simulation and real-world datasets and providing the open-source code for further development.
\end{enumerate}
This paper is organized as follows.
Section II introduces the related work.
Section III gives an overview of the Phase-SLAM system pipeline.
Section IV describes the proposed phase-based pose estimation and compressive loop detection methods. 
Section V provides experiment results and discussions. 
Section VI concludes the paper and outline the future works. 
Appendix supplements the details of the Jacobian matrix in use.
\section{Related work}	

Most visual SLAM systems are based on either direct or indirect schemes. Direct approaches \cite{lsd14, engel2017direct} sample pixels from image regions and minimize the photometric error. Indirect approaches \cite{mur2015orb, qin2018vins} require extra computational resources for detecting and describing features.
In contrast, the proposed Phase-SLAM system is based on pixel-level phase data, which contain 3D depth information and can be extracted directly by selecting a region of interest (ROI).

\subsection{Point Cloud Registration}
SLI systems often use point cloud registration methods to achieve large fields of view scans, either local or global.
Classical local registration methods, such as Point-to-Point ICP \cite{ICP1992}, minimizes the sum of distances between points and their nearest neighbours.
Point-to-Plane ICP \cite{low2004linear} assumes that each corresponding point is located on a plane, and introduces surface normals into the objective function to achieve more efficient data registration.
Symmetrized objective function (SymICP) have been proposed to extend the planar convergence into a quadratic one at extra computational costs \cite{rusinkiewicz2019symmetric}. 

Local methods are limited by initial guesses, so structural features of point clouds are used to search for transformations globally.
Point coordinates and surface normals have been used to compute the Fast Point Feature Histograms (FPFH) \cite{FPFH2009}, and the coplanar 4-point sets have been chosen as features for registration (Super 4PCS) \cite{mellado2014super}.
Besides,
Go-ICP uses the branch-and-bound (BnB) scheme to avoid local optima \cite{yang2015go}. 
Fast Global Registration (FGR) applies a Black-Rangarajan duality to achieve a more robust objective function \cite{zhou2016fast}. 
BCPD++ formulates coherent point drift in a Bayesian setting to supervise the convergence of algorithm \cite{BCPD++}.
Compared with above 3D point cloud registration methods, our approach converts 3D point cloud registration into 2D phase data registration, resulting in much reduced computational complexity and memory footprint.

\subsection{Loop Closure Detection}

Loop closure detection can effectively eliminate the accumulating error. 
A plain method is randomly sampling a number of keyframes to find loop closures \cite{endres20133}.
Odometry based approaches judge whether there is a loop closure at the current position according to the calculated map \cite{hahnel2003efficient}.
Appearance based approaches determine the loop relationship based on the similarity of two scenes \cite{mur2015orb, qin2018vins}. 
Bag-of-Words (BoW) based the approach \cite{galvez2012bags} uses descriptors (words) for loop closure detection instead of whole images.
In this paper, our loop closure detection is based on compressed phase data to reduce both computatonal complexity and storage space without losing much detection performance.


\section{System Setup and Problem Statement}
\begin{figure*}
	\centering
	\vspace{-0.3cm}
	\setlength{\abovecaptionskip}{-0pt}
	\includegraphics[width=0.999\textwidth]{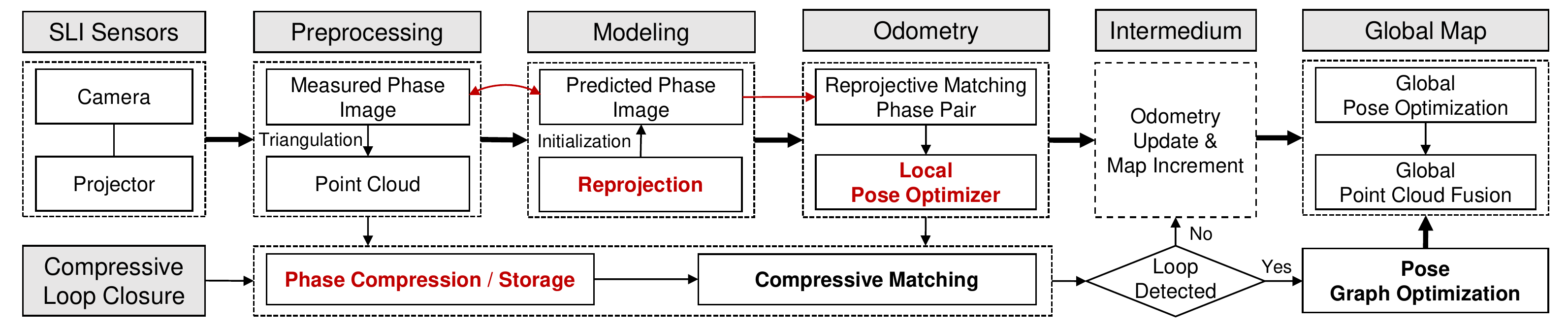}
	\caption{The system diagram of the proposed Phase-SLAM. 
		Based on the SLI phase image, 3D point clouds for each new sensor pose can be computed,
		which are used for phase data prediction (Section IV-B) and local pose estimation (Section IV-C). 
		The compressive loop closure detection is performed to trigger the global pose graph optimizer (Section IV-E). 
		Finally, the refined sensor poses are used to achieve 360 degree 3D point clouds of the object under scanning.}
	\label{fig: sys workflow}
\end{figure*}

\subsection{System Setup}

The proposed Phase-SLAM pipeline is shown in Fig. \ref{fig: sys workflow}. Based on the SLI sensor data, correponding 3D point clouds and the reprojection model are used to obtain phase data (Section IV-B).
Then, the local pose optimization module (Section IV-C) is used to estimate sensor poses by minimizing errors between predictions and measurements of phase data. 
Local pose graphs are updated until the compressive loop closure detection (Section IV-E) is triggerred. 
The pose graph optimizer then performs global optimization to eliminate the cumulative errors and revise sensor poses. 
Finally, poses are used to align multi-view point clouds and achieve the overall 3D object reconstruction.

We define the notations used in this paper. 
The initial position of the projector is chosen as the origin of the world coordinate system. 
$(\cdot)^w$ is the world frame, $(\cdot)^c$ is the camera frame, $(\cdot)^p$ is the projector frame, and $(\cdot)_k$ means the $k$-th sensor pose. The $\Phi$ and $\phi$ stand for the phase image and phase value at each pixel location, respectively. $\hat{(\cdot)}$ denotes the estimated value. $\mathbf P(x, y, z)$ is the 3D coordinate of a point. The transformation between two sensor poses is represented by vector $\Delta \mathbf{X}=[\delta x, \delta y, \delta z, \delta \alpha, \delta \beta, \delta \gamma]$,  Matrix $\mathbf{R}$ and vector $\mathbf{t}$ represent rotation and translation from $pose_k$ to $pose_{k+1}$. $\mathbf{R}$ and $\mathbf{t}$ can be obtained for a given $\Delta \mathbf{X}$.


\subsection{Problem Statement}
This work aims at developing a complete SLAM system that can estimate the SLI sensor pose transformation $\Delta \mathbf{X}$ through phase data registration
and achieve global 360 degree dense 3D reconstruction through pose graph optimization. 
At each step, 3D points are projected into the sensor imaging plane with initialized pose rotation and translation $\mathbf{R}_k, \mathbf{t}_k$ by using
\begin{equation} \label{reproj}
	\mathbf{u}_{k+1}(\mu, \nu) = \pi_{\mathcal{M}}(\mathbf{R}_k\mathbf{P}^w_k + \mathbf{t}_k),
\end{equation}
where the $\pi_{\mathcal{M}}$ is the perspective transformation with the projection matrix $\mathcal{M}$, means $\mathbb{R}^3 \rightarrow \mathbb{R}^2$ that projects a 3D point onto the imaging plane. $\mathbf{u}_{k+1}$ is the pixel position, which is used for obtaining phase data estimations $\hat{\phi}$ and measurements $\phi$.
Obtained the $\hat{\phi}$ and $\phi$, the sensor pose transformation $\Delta \mathbf{X}$ is estimated by
\begin{equation} \label{Cfunction}
		\Delta \mathbf{X}^{*} = \arg \min \limits_{\Delta \mathbf{X}} \mathbf{F}(\Delta \mathbf{X}),
\end{equation}
where 
\begin{equation}
	\mathbf{F}(\Delta \mathbf{X}) = 
	\frac{1}{2}\sum
	{\left\| 
		\mathbf{\hat{\phi} - \phi} \right\|^2}.
\end{equation}
Such a local optimization procedure requires computing the Jacobian matrix iteratively until it converges. 
The loop closure detection and pose graph optimization will be also needed to reduce estimation errors.

\subsection{SLI Scanning} 
\begin{figure}
	\centering
		\vspace{-0.2cm}
	\setlength{\abovecaptionskip}{-2pt}
	\includegraphics[width=0.49\textwidth]{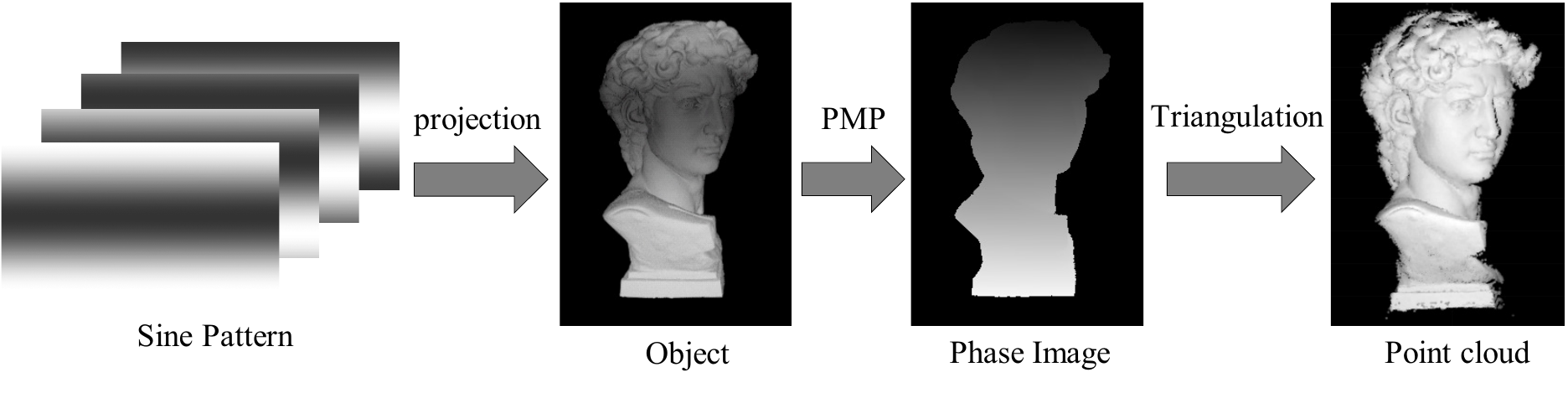}
	\caption{An illustration of SLI imaging system. 
		PMP uses the images of projection patterns to compute phase images; the 3D point clouds are then obtained 
		by triangulation with the calibrated camera-projector parameters.}
	\label{fig: SLI workflow}
\end{figure}

In the camera-projector based SLI system, the Phase Measuring Profilometry (PMP) method is used to calculate the phase image, as shown in Fig. \ref{fig: SLI workflow}.
The camera captures the raw images of sine patterns deformed by the scanned surface, given by
\begin{equation} \label{eq:PMP_1}
	I_n^c\left( {{\mu},{\nu}} \right) = {A} + {B}\cos\left( {\Phi \left( {{\mu},{\nu}} \right) - \frac{{2\pi n}}{N}} \right),
\end{equation}
where 
$n = 1, 2, \cdots N$ (the number of patterns), $A$ and $B$ are the background brightness and intensity modulation, respectively. The phase image $\Phi \left( {{\mu},{\upsilon}} \right)$ can then be calculated by \cite{wang2012robust}
\begin{equation} \label{eq:PMP_2}
	\Phi {\rm{ = arctan}}\left[ {\frac{{\sum\nolimits_{n = 1}^{N} {I_n^c} \sin \left( {{{2\pi n} \mathord{\left/
							{\vphantom {{2\pi n} N}} \right.
							\kern-\nulldelimiterspace} N}} \right)}}{{\sum\nolimits_{n = 1}^{N} {I_n^c} \cos \left( {{{2\pi n} \mathord{\left/
							{\vphantom {{2\pi n} N}} \right.
							\kern-\nulldelimiterspace} N}} \right)}}} \right].
\end{equation}
\section{Proposed methods}

This section investigates the geometric model among 3D point, phase data and sensor pose. 
Comparing with our previous work \cite{zheng2021phase}, this work develops a more intuitive and simpler model based on reprojective transformation method. 
After phase data pairing, the sensor pose motion can be estimated through least-square optimization between phase predictions and measurements. 
A compressed sensing scheme is adopted to achieve fast loop closure detection for global pose graph optimization.
\subsection{Phase Values under Epipoloar Constraint}

\begin{figure}
	\centering
	\vspace{-0.3cm}
	\setlength{\abovecaptionskip}{-2pt}
	\includegraphics[width=0.48\textwidth]{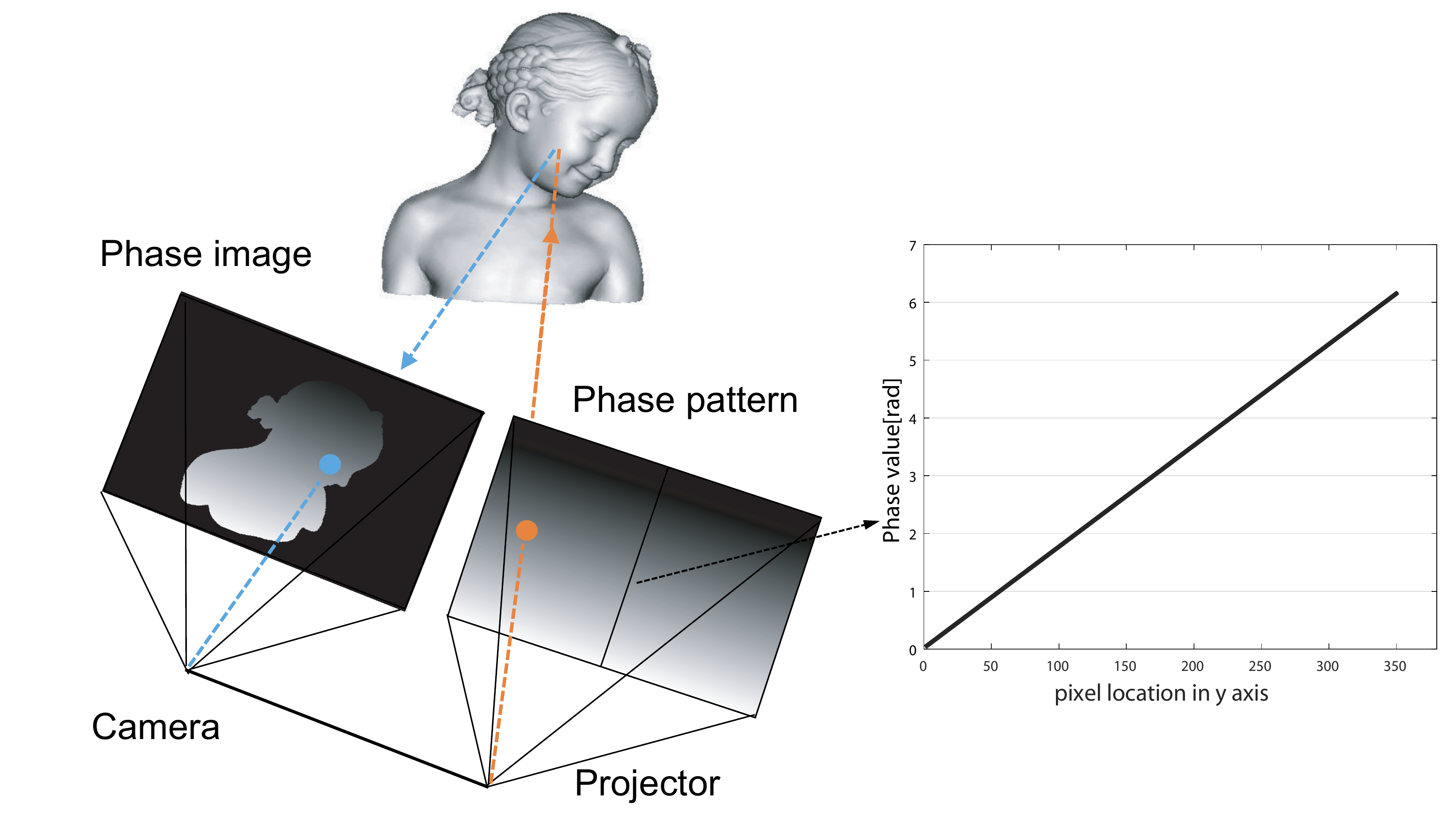}
	\caption{An illustration of SLI imaging principle. 
	The projector is regarded as a camera with the phase pattern. 
	The phase pair should be performed under the epipoloar constraint.
	The phase values of each column of phase pattern are linearly increased from 0 to 2$\pi$. }
	\label{fig: pro2cam}
\end{figure}

In a SLI system, we regard the projector as another camera, which has similar projection parameters and perspective principles with it. 
As shown in the left of Fig. \ref{fig: pro2cam},
according to the epipoloar constraint, a phase value obtained from the phase image can correspond to a pixel location in the ``camera" imaging plane (phase pattern), like stereo-vision \cite{andrew2001multiple}. 
In PMP method, the phase pattern is actively projected by the projector, so pixel locations and phase values on pattern plane have a fixed and known relevance. 
As shown in the right of Fig. \ref{fig: pro2cam}, the phase value of each column in phase pattern is linearly increased from 0 to 2$\pi$ and each row in the pattern is the same. 
This means when we get the phase value of a 3D point $\mathbf{P}$ from the phase image, we can know its ordinate under the projector's pattern coordinate. Vice versa, if we knew the projective coordinate $({\mu^p},{\nu^p})$ (just only ${\nu^p}$) of a point $\mathbf{P}$ in the pattern, we could get its corresponding phase value on the phase image by 
\begin{equation}\label{phi}
	\phi = 2 \pi \nu^p/H^p ,
\end{equation}
where $H^p$ is the row height of the projector's imaging plane. 

\subsection{Phase Pairing Based on Reprojective Transformation}



\begin{figure}
	\centering
	\vspace{-0.4cm}
	\setlength{\abovecaptionskip}{-2pt}
	\includegraphics[width=0.5\textwidth]{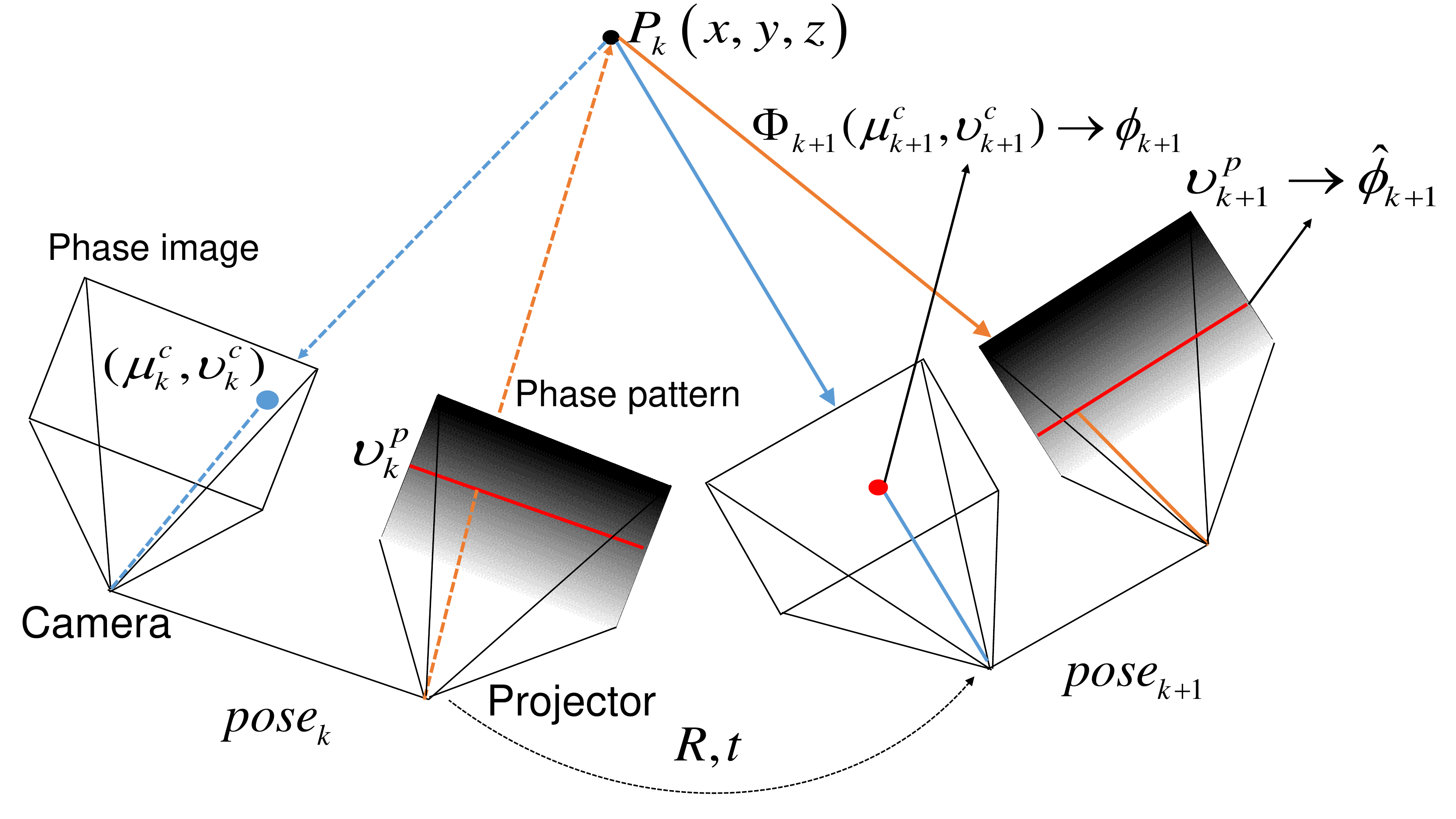}
	\caption{An illustration of reprojection from 3D points to 2D phase data. A 3D point P obtained by $pose_k$ is reprojected into the imaging plane of camera and projector in $pose_{k+1}$ with a assumed rotation $\mathbf{R}$ and translation $\mathbf{t}$. Then the errors between the predicted and measured phase data ($\hat{\phi}_{k+1}$ and $\phi_{k+1}$) are minimized with respect to $\mathbf{R}$ and $\mathbf{t}$.}
	\label{fig: projection}
\end{figure}

As shown in the Fig. \ref{fig: projection}, a 3D point $\mathbf{P}$ is measured by the SLI sensor at the $pose_k$  with the coordinate $\mathbf{P}_k=[x, y, z]^\top$. Assuming the transformation: rotation matrix $\mathbf{R}(\delta \alpha, \delta \beta, \delta \gamma) \in SO(3)$ and translation vector $\mathbf{t} = [\delta x, \delta y, \delta z]^\top$, the SLI move to $pose_{k+1}$ by it and the point $\mathbf{P}$ will have a new coordinate $\mathbf{P}_{k+1} = [x', y', z']^\top$, given by
\begin{equation} \label{transform}
	\mathbf{P}_{k+1} = \mathbf{R}\mathbf{P}_k + \mathbf{t}.
\end{equation}
Based on the new coordinate, the point $\mathbf{P}$ is reprojected into camera and projector imaging plane in $pose_{k+1}$ to get two pixel locations on them: $\mathbf{u}^c_{k+1}(\mu^c_{k+1}, \nu^c_{k+1})$ and $\mathbf{u}^p_{k+1}(\mu^p_{k+1}, \nu^p_{k+1})$ by the transformation $\pi_{\mathcal{M}}$ (Eq. (\ref{reproj})), respectively.
On the projector imaging plane (the phase pattern in Fig. \ref{fig: pro2cam}),
when the reprojection ordinate of point $\mathbf{P}$: $\nu^p_{k+1}$ is known, the phase value prediction $\hat{\phi}_{k+1}$ can be obtained by Eq. (\ref{phi}). 
So, combining Eq. (\ref{reproj}, \ref{phi}, \ref{transform}), the phase value of the a 3D point in the phase pattern can be estimated by using
\begin{equation} \label{eq:3dof}
	\hat{\phi}_{k+1} = 
	\frac{2\pi}{H^p} {\left(\frac{ f^p_y(R_{21}x + R_{22}y + R_{23}z + \delta y)}
		{R_{31}x + R_{32}y + R_{33}z + \delta z}
		+ C^{p}_{y} \right)},
\end{equation}
where  $R_{ij}$ is the $ij^{th}$ element of $\mathbf{R}$, $f^p_y$, $C^{p}_{y}$ is calibration parameter (projector's focal length and principal point along the row of the projector imaging plane, respectively). 

On the camera imaging plane, the phase value measurement $\phi_{k+1}$ can be obtained from the phase image $\Phi_{k+1}$ at the pixel location $({\mu^c_{k+1}},{\nu^c_{k+1}})$, which can be computed by
\begin{align}\label{eq:m_k+1}
	\begin{split}
		\mu^c_{k+1} &= \frac{m_{11}x' + m_{12}y' + m_{13}z'}
		{m_{31}x' + m_{32}y' + m_{33}z' } \\
		\nu^c_{k+1} &= 
		\frac{ m_{21}x' + m_{22}y' + m_{23}z'}{m_{31} x' + m_{32} y' + m_{33} z'},
	\end{split}
\end{align}
where $m_{ij}$ is the $ij^{th}$ element of projection matrix $\mathcal{M}$.
When ${\mu^c_{k+1}}$ and ${\nu^c_{k+1}}$ are not integers, bilinear interpolation on $\Phi_{k+1}$ can be used to calculate phase data at integer indices. 

\begin{figure}
	\centering
	\vspace{-0.4cm}
	\setlength{\abovecaptionskip}{-2pt}
	\includegraphics[width=0.45\textwidth]{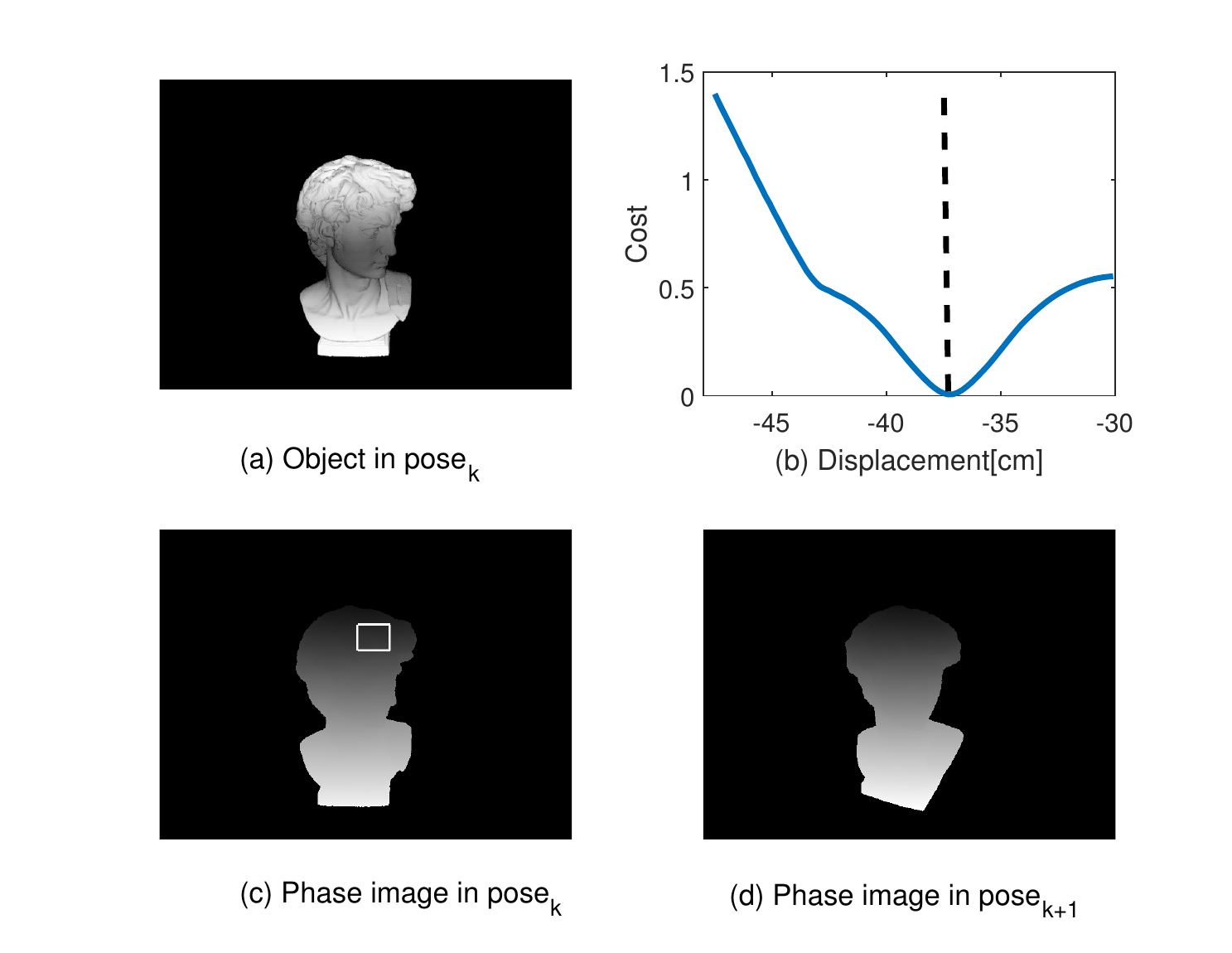}
	\caption{An illustration of a simple local pose optimization process. (a) an object raw image acquired at $pose_k$;
	(c) the phase image acquired at $pose_k$ with a white ROI; 
	(d) the phase image acquired at $pose_{k+1}$;
	(b) the plot of optimization objective cost between two phase images (within the ROI) with respect to different sensor displacements, where the black dashed line indicates the ground truth of the minimum.}
	\label{fig: convergence}
\end{figure} \vspace{-0.2cm}

\subsection{Local Pose Optimizer}
In the local optimizer, the state variables are defined as $\Delta \mathbf{X}(\delta x, \delta y, \delta z, \delta \alpha, \delta \beta, \delta \gamma)$, which is equivalent to $\mathbf{R}$ and $\mathbf{t}$.
The error $\mathbf{e}$ between $\hat{\phi}_{k+1}$ and  $\phi_{k+1}$ are given by
\begin{equation}\label{e}
	\mathbf{e} = \hat{\phi}_{k+1}(\Delta \mathbf{X}) -\Phi_{k+1}(\mu_{k+1}(\Delta \mathbf{X}),\nu_{k+1}(\Delta \mathbf{X})).
\end{equation}
The objective function is shown in           
\begin{equation} \label{eq:F_function}
		 \mathbf{F}(\Delta \mathbf{X}) = 
		\frac{1}{2}\sum_{\langle \mu, \nu \rangle \in \mathbb{R}} 
	    {\left\| 
	  \mathbf{\mathbf{e}} \right\|^2},
\end{equation}
the $\mathbb{R}$ is a ROI in phase images, $ (\cdot)^i$ is the $i$-th point in ROI. $\mathbf e = [\mathbf e^1, \mathbf e^2, \cdots, \mathbf e^n]^\top$.

The proposed function Eq. (\ref{eq:F_function}) can be solved by iterative gradient-based methods \cite{nocedal2006numerical}.
Given the initial value $\Delta {\widetilde{\mathbf X}}$, the cost function can be approximated by Taylor expand about $\Delta {\widetilde{\mathbf X}}$, and $\mathbf F(\Delta {\widetilde{\mathbf X}} + \bm \Delta) \approx \mathbf F(\Delta {\widetilde{\mathbf X}}) +  \nabla \mathbf F \bm \Delta$, where
\begin{equation}\label{J1}
	\begin{split}
		\nabla \mathbf F &= \mathbf{J^\top e} \\
		\mathbf{J} &= \partial{\mathbf{e}} / \partial{\Delta \mathbf{X}}.
	\end{split}	
\end{equation}
$\mathbf J$ is the Jacobian matrix, the optimization increment $\bm \Delta$ is computed by $\lambda \bm \Delta = -\mathbf{J^\top e} $, which is the negative gradient direction of $\mathbf{F}$, $\lambda$ controls the size of steps. The solution is updated by  $\Delta \mathbf{X}_{i+1} = \Delta \mathbf{X}_{i} + \bm{\Delta}_{i}$, where
$i$ is the iterative index \cite{andrew2001multiple, nocedal2006numerical}.

Fig. \ref{fig: convergence} shows a simple example the optimization process. 
Fig. \ref{fig: convergence} (a) shows an object image. 
(c) shows the corresponding phase image with a ROI.
(d) shows the phase images acquired at a new sensor pose. 
(b) shows the plot of errors between two sets of phase data (within the ROI) with respect to the displacements of the SLI sensor.
It can be seen that such an optimization process can be converged to the local minimum \cite{zhou2018semi}.

\subsection{The Jacobian Matrix}
According to Eq. (\ref{e}, \ref{J1}), the Jacobian matrix of $\mathbf e^i$ ($i = 1, 2, \cdots, n$) is given by
\begin{equation} \label{eq:J}
	\begin{split}
		\mathbf J^{i}  = \frac{\partial \mathbf e^i} {\partial \Delta \mathbf X} 
		= \frac{\partial \hat \phi^i_{k+1}} {\partial \Delta \mathbf X}
		- \left(\frac{\partial \Phi_{k+1}} {\partial \mu^i_{k+1}}
		\frac{\partial \mu^i_{k+1}} {\partial \Delta \mathbf X}
		+ \frac{\partial \Phi_{k+1}} {\partial \nu^i_{k+1}}
		\frac{\partial \nu^i_{k+1}} {\partial \Delta \mathbf X}\right)
	\end{split},
\end{equation}
where
$\displaystyle{{\partial \Phi_{k+1}}/{\partial \mu_{k+1}}}$ and $ \displaystyle{{\partial \Phi_{k+1}}/{\partial \nu_{k+1}}}$ are vertical and horizontal gradients of $\Phi_{k+1}$, computed by pixel difference. 
More details in Eq. (\ref{eq:J}) are provided in the Appendix, and
the other term in the Eq. (\ref{J1}) is substituted by
\begin{equation}\label{J_3}
	\mathbf{J^\top e} = \sum_{i = 1}^{n} \mathbf{e}^i \displaystyle{\frac{\partial \mathbf{e}^i}{\partial \Delta \mathbf X}} .
\end{equation}

\subsection{Loop Closure Detection}

\begin{figure}
	\centering
	\vspace{-0.0cm}
	\setlength{\abovecaptionskip}{-2pt}
	\includegraphics[width=0.45\textwidth]{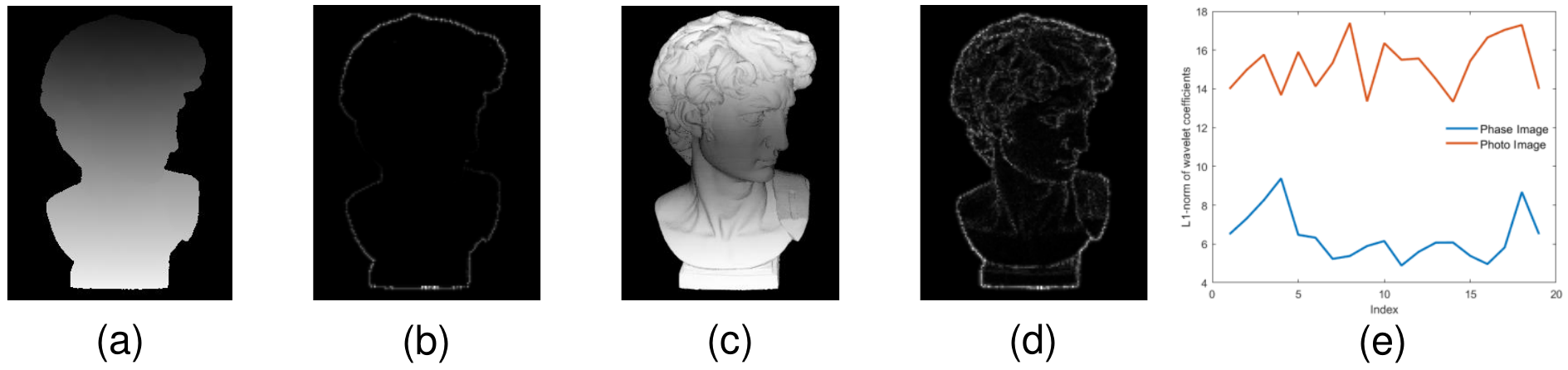} 
	\caption{An illustration of the sparsity of phase and photo images. (a,c) A phase image and a photo image; (b,d) the corresponding wavelet coefficients of two images; (e) the wavelet coefficient L1 norms of two types of images within one SLAM loop. }
	\label{fig: sparsity}
\end{figure} \vspace{-0.0cm}

The proposed Phase-SLAM utilizes the Compressive Sensing (CS) technique to reduce computational complexity and data storage space for loop closure detection. The compressibility of an image is determined by its sparsity. More sparse images will lose less information after compression and the sparse image contains less high-frequency information \cite{candes2008restricted}. 
A haar wavelet bases and $L_1$ norm are used to illustrate the degree of sparsity of phase images. As shown in Fig. \ref{fig: sparsity}, a phase image and a photo image are projected upon wavelet bases first. Then the $L_1$ norms of the wavelet coefficients of two types of images within one SLAM loop are compared in Fig. \ref{fig: sparsity} (e). It can be seen that the $L_1$ norms of the wavelet coefficients of phase images are much smaller than photo images, indicating the degree of sparsity of phase images is much smaller than photo images.

According to the CS theory \cite{candes2008restricted},  
two signals ${A_1}$ and ${A_2}$ are distinguishable after compression if the matrix $\mathbf{C}$ satisfies
\begin{equation} \label{eq:rip}
	2(1-\delta_{2s}) \le ||\mathbf{C} {A}_2 - \mathbf{C} {A}_1||^2_{2} \le 2(1+\delta_{2s}),
\end{equation}
where $\delta_{2s}$ is a constant, and $\mathbf{C}$ is Gaussian matrix. 
The compressed signals ${y}_{n\times1}=\mathbf{C}_{n\times N} {A}_{N\times1}$  $n<N$, has quite smaller size than the original signals.
For a 2D phase image $\Phi$, we first reshape it into a 1D vector $\Phi'$. 
The reshaped phase data $\Phi'$ can be recovered by the compressed signal $ {y} = \mathbf{C} \Phi'$,
and the error between two compressive phase vector is shown in
\begin{equation} \label{eq:cs-loop}
	dy = ||\mathbf{C} \Phi'_2 - \mathbf{C} \Phi'_1||^2_{2} .
\end{equation}
When $dy$ is smaller than a threshold, the loop-closure is detected. 


\subsection{The Pipeline of Phase-SLAM}
After successful loop-closure detection, the pose graph optimization technique \cite{grisetti2010tutorial} will be used to eliminates cumulative error and refine poses. 
The pose sequences in our system usually have a large interval during the scanning process, so every estimated pose is a vertex in the pose graph optimizer. 

\section{EXPERIMENT RESULTS AND DISCUSSIONS}

The proposed Phase-SLAM system was evaluated with both the Unreal Engine 4 (UE4) simulator and real-world experiments. 
All experiments were implemented on a PC with an Intel Core i7-9800K CPU @ 3.6GHz.

\subsection{Simulation Experiments}

\begin{figure}
	\centering
	\vspace{-0.0cm}
	\setlength{\abovecaptionskip}{-2pt}
	\includegraphics[width=0.48\textwidth]{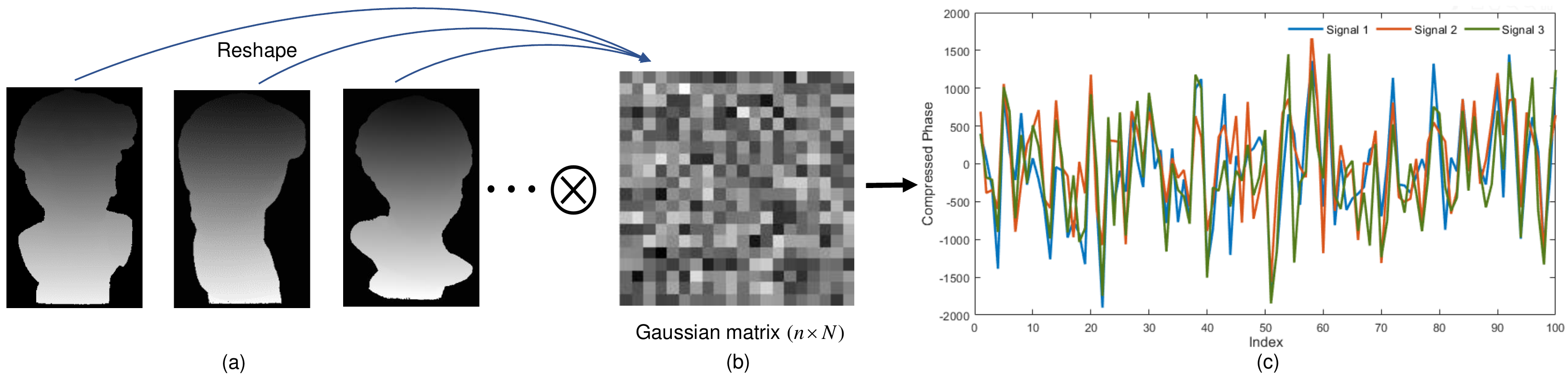} 
	\caption{An illustration of the proposed compressive loop closure detection. (a) Three phase images; (b) the Gaussian pseudo-random matrix used for compressive projection; (c) three sets of compressed signals corresponding to three phase images used for loop closure detection.}
	\label{fig: compressphase}
\end{figure} \vspace{-0.cm}
\begin{figure*}
	\centering 
	\vspace{-0.4cm}  
	\setlength{\abovecaptionskip}{-0pt}
	\includegraphics[width=0.85\textwidth]{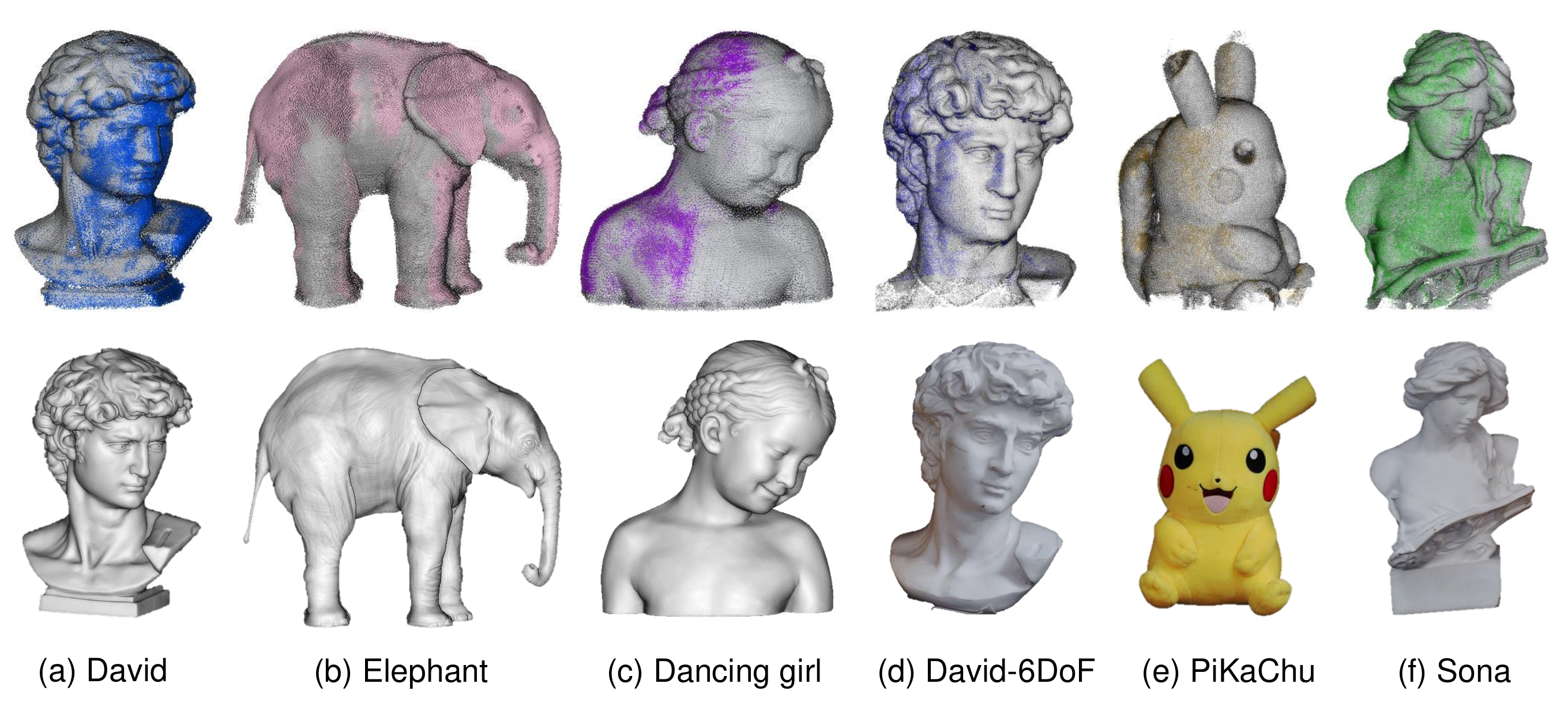}
	\caption{A comparison of global 3D point cloud registration results along with ground-truth. (a)-(c) Simulation datasets named David, Elephant and Dancing girl; (e)-(f) real-world datasets named David-6DoF, PiKaChu and Sona. (Top Row) The ground truth (in gray) and reconstruction results (in other colors) by using the proposed Phase-SLAM; (Bottom Row) the 3D objects.}
	\label{fig: PCall}
\end{figure*}
The simulation dataset was collected with the Airsim plugin in UE4. Different 3D models were used as targets, and the virtual SLI device moved around the target along a radius of 120 cm and with a rotation interval of 20 degrees.  
The simulated dataset is based on three models namely David, Elephant and Dancing girl, which contains calibration parameters, phase images and ground-truth poses.
The baseline methods include four state-of-the-art local methods, namely Point-to-Point ICP \cite{ICP1992}, Point-to-Plant ICP \cite{low2004linear}, SymICP \cite{rusinkiewicz2019symmetric} and FPFH \cite{FPFH2009}, 
and two SOTA global methods, namely FGR \cite{zhou2016fast} and BCPD++ \cite{BCPD++}.
Local methods were conducted based on Point Cloud Library (PCL) implementation \cite{PCL}. Global methods were based on open-source code. 
The numbers of iterations of Point-to-Point ICP, Point-to-Plane ICP, and SymICP were chosen as 30; FPFH was 10000. 
The implementation of FGR and BCPD++ used the recommended parameters.
\begin{figure}
	\centering
	\vspace{-0.0cm}
	\setlength{\abovecaptionskip}{-2pt}
	\includegraphics[width=0.42\textwidth]{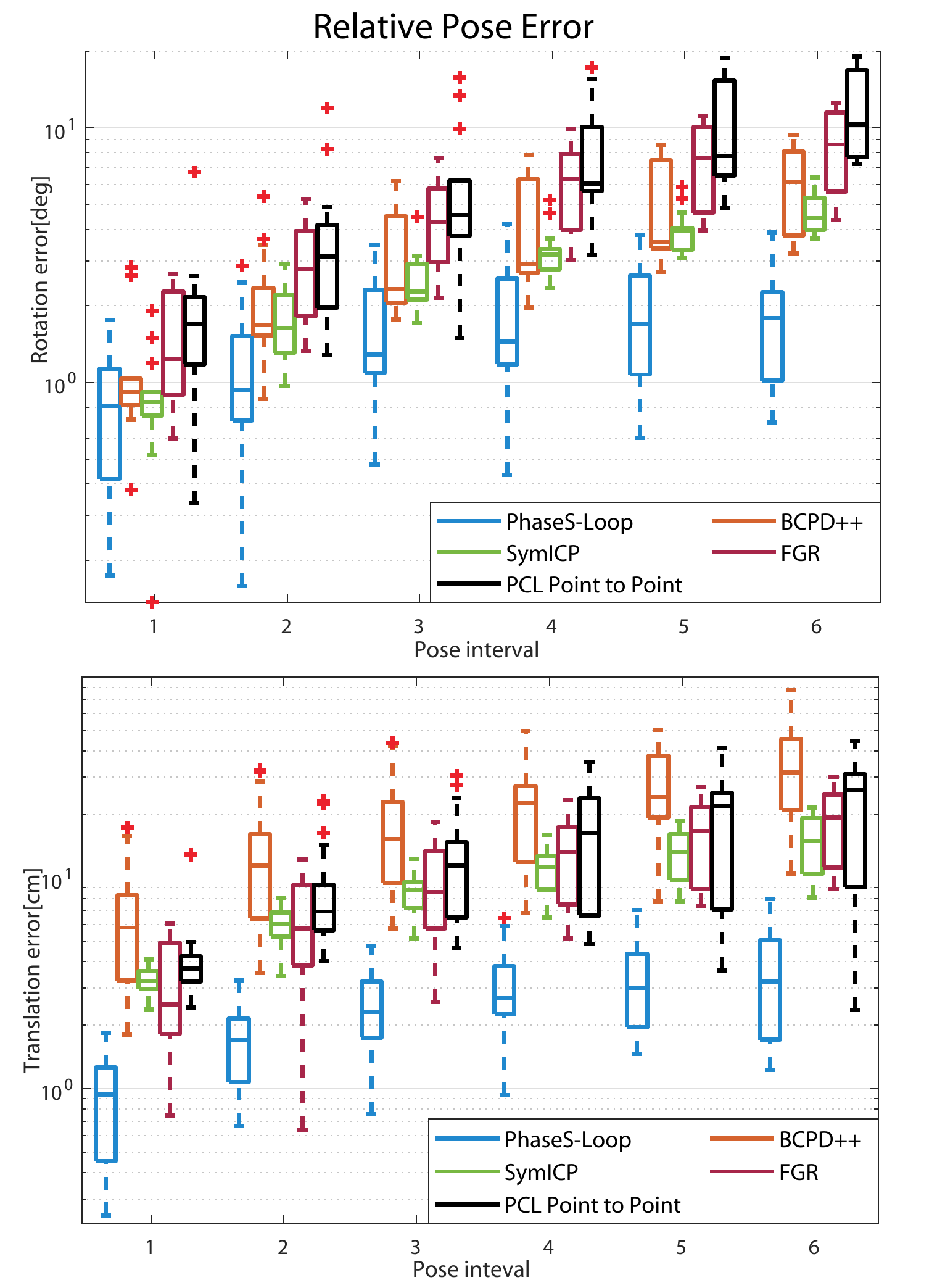}
	\caption{The plot of Relative Pose Errors \cite{ATE} of (top) rotation and (bottom) translation in simulation. PhaseS-Loop denotes Phase-SLAM with pose graph optimization.}
	\label{fig: RPE}
\end{figure} \vspace{-0.0cm}

The compression of phase images is shown in Fig. \ref{fig: compressphase}. 
In simulation experiments, the resolution of phase images is $640\times 480$ (Fig. \ref{fig: compressphase} (a)),
the size of Gaussian compressive random matrix is chosen as $100 \times 307000$ (Fig. \ref{fig: compressphase} (b)),
that is, the compression ratio is 3070:1 and the size of the compressed phase signal is $100 \times 1$. 
Fig. \ref{fig: compressphase} (c) shows the compressed signals corresponding to three phase images like Fig. \ref{fig: compressphase} (a). 
It can be seen that the three sets of signals are distinguishable in terms of the peaks and valleys for loop closure detection.
Furthermore, experiment results show that the time consumption of the back-end optimization using CS technique can be reduced by $20\%$ than using original phase images.

Fig. \ref{fig: PCall} (a)-(c) shows the 3D reconstruction results (top) and ground truth (bottom) for the three simulation targets (David, Elephant and Dancing girl) using the proposed Phase-SLAM with loop closure.
More quantified reconstruction errors are shown in Fig. \ref{fig:star}.
Fig. \ref{fig: RPE} shows the relative pose errors (RPE) \cite{ATE} in rotation (top) and translation (bottom), respectively by using five methods with David dataset.
It can be seen that the proposed Phase-SLAM method with loop closure (PhaseS-Loop) outperforms other four methods.
The median RPE of Phase-SLAM is 0.81 degree and 0.94cm in rotation and translation, respectively.
Table \ref{tab:RMSEunreal} shows the root mean squared error (RMSE) of absolute trajectory error (ATE) \cite{ATE} and the computation time for three different datasets.
The average RMSE of our approach is 1.06cm, which is better than other methods. 
Actually, Phase-SLAM with loop closure outperforms PhaseS by $38.5\%$. 
In simulations, the average number of 3D points corresponding to the image is around 50000. 
BCPD++ has the highest compuation speed among the 6 existing methods. 
Our approach is still almost two times faster than BCPD++. 
And the average running time of the back-end optimization is 40.7 ms.  
\begin{table} \vspace{-0.2cm}
	\centering
	\caption{RMSE of ATE (cm) / Computation time (s)}
	\begin{tabular}{c c c c}  
		\toprule[1.5pt]
		Method         & David          & Elephant       & Dancing Girl \\
		\midrule[1pt]
		PhaseS-Loop[ours]    & \textbf{1.39 / 1.52} & \textbf{0.72 / 1.58} & \textbf{1.07 / 0.82} \\
		PhaseS[ours]         & \textbf{2.32 / 1.49} & \textbf{2.40 / 1.23} & \textbf{2.05 / 0.76} \\
		BCPD++\cite{BCPD++}         & 25.53 / 2.87        & 87.09 / 3.46        & 15.57 / 2.68 \\
		SymICP\cite{rusinkiewicz2019symmetric}       & 6.35 / 109.10       & 7.06 / 146.71   & 8.17 / 104.22\\
		Point to Plane\cite{low2004linear} & 6.76 / 55.15        & 13.07 / 65.73       & 10.57 / 45.89\\
		Point to Point\cite{ICP1992} & 17.17 / 32.56      & 19.35 / 44.68      & 40.26 / 30.75\\
		FGR\cite{zhou2016fast}       & 11.12 / 25.56         & 15.67 / 34.68    & 38.74  / 26.90\\
		FPFH\cite{FPFH2009}          & 8.99 / 70.59         & 25.10 / 79.09        & 32.70 / 59.68\\  
		\bottomrule[1.5pt]
	\end{tabular}                         
	\label{tab:RMSEunreal}
	\begin{tablenotes}
		\item[1] RMSE of ATE: The root mean squared error of absolute trajectory error.
	\end{tablenotes}
\end{table} \vspace{-0.0cm}

\subsection{Real-World Experiments}

\begin{figure}
	\centering
	\vspace{-0.0cm}
	\includegraphics[width=0.48\textwidth]{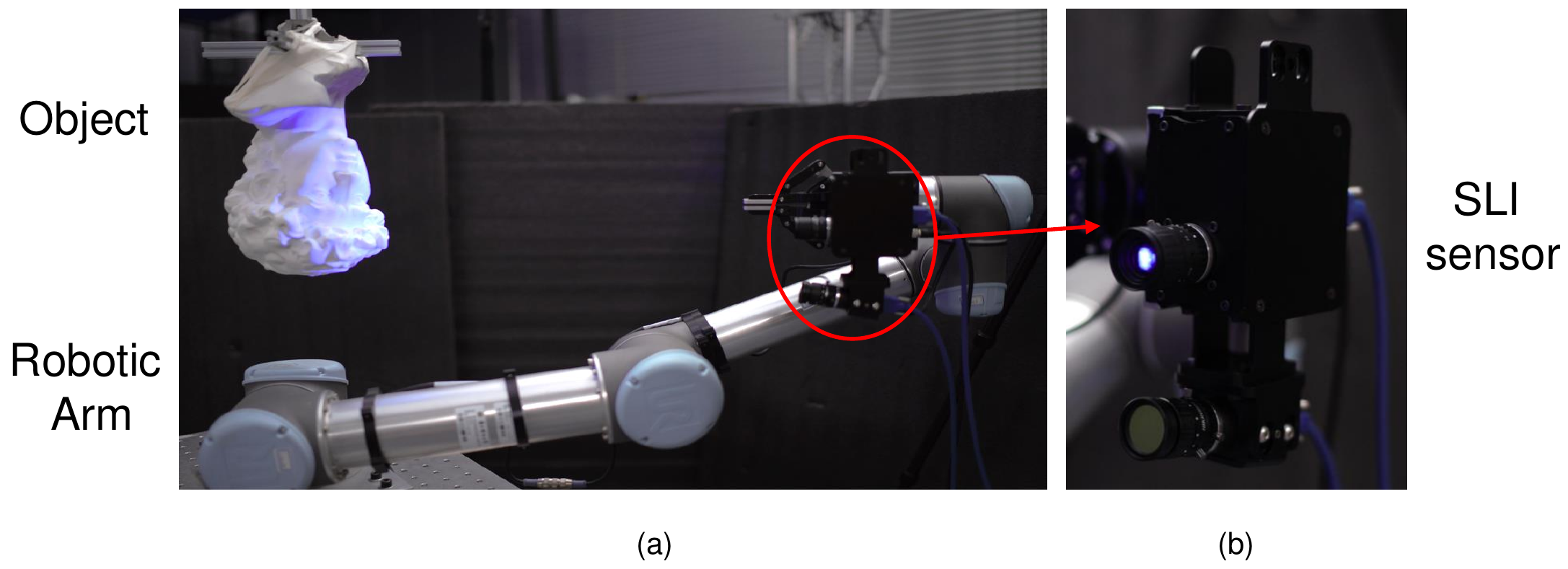}
	\caption{(a)The real-world experiment setup where a object is fixed on a bracket and the SLI sensor is installed on a UR5 robotic arm. (b)The SLI sensor consists of a DLP3000 projector and a HIKVISION camera.}
	\label{fig:SLI sensor} \vspace{-0.2cm}
\end{figure}

Fig. \ref{fig:SLI sensor} shows the experiment setup, where
the SLI sensor, consisting of a projector (DLP3000 DMD from TI) and an industrial camera (1280$\times$1024 resolution from HIKVISION), is mounted on a UR5 robotic arm. 
Fig.  \ref{fig: PCall} (d-f) show two plaster statues (David, Sona) and a plush toy (PiKaChu) used to build real-world datasets, 
namely David-6DoF, David-3DoF, Sona-3DoF and PiKaChu-3DoF, where 6DoF and 3DoF stand for six and three degrees of freedom motions, respectively. 
The David-6DoF dataset includes 31 random poses; Each 3DoF dataset has 37 poses at equal intervals of 10 degrees and a radius of 60cm.

\begin{figure}
	\centering
	\vspace{-0.4cm}
	\includegraphics[width=0.42\textwidth]{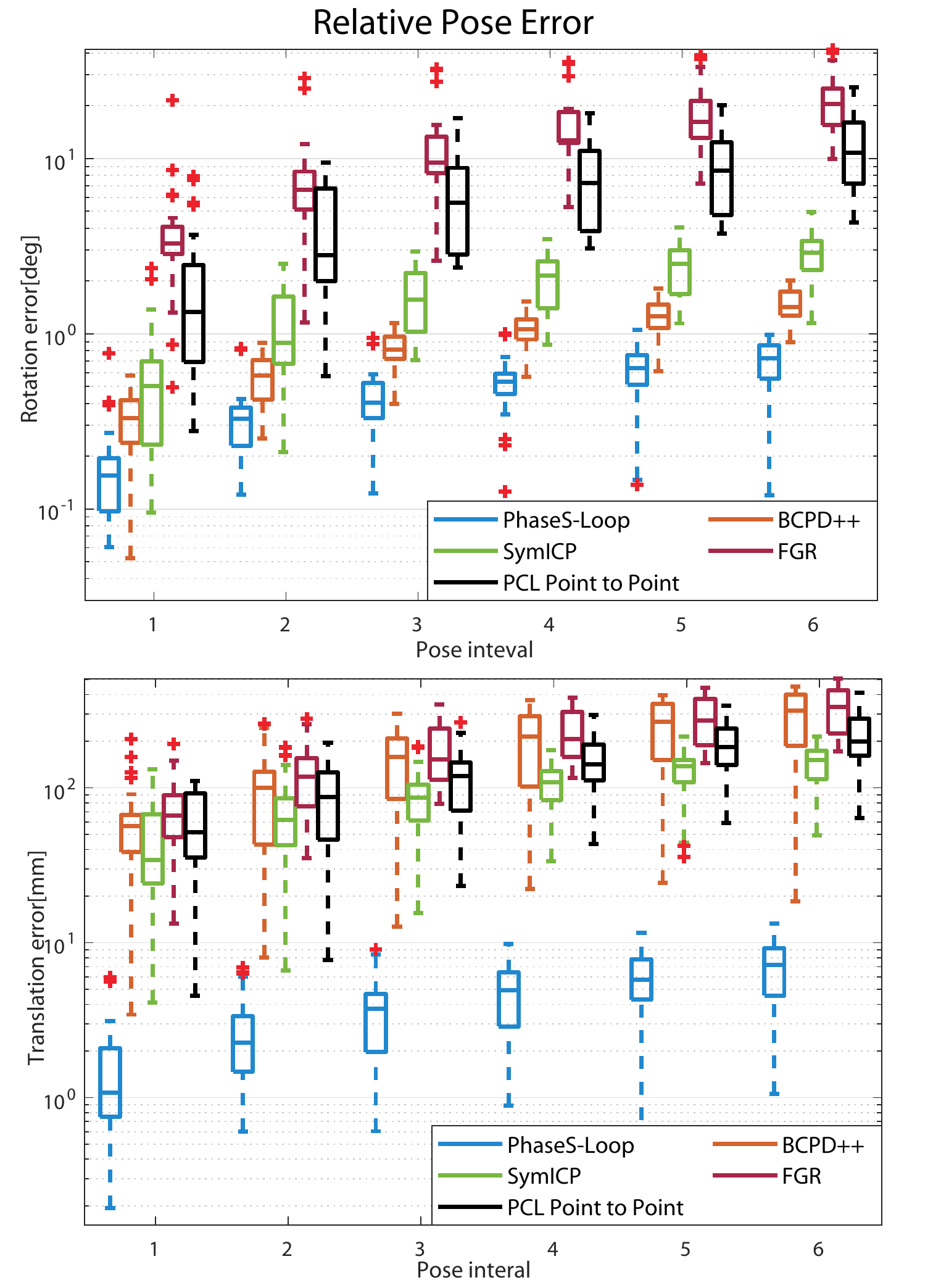}
	\caption{The plot of Relative Pose Errors \cite{ATE} of (top) rotation and (bottom) translation in real-world experiments.}
	\label{fig: RPEReal}
\end{figure}

\begin{figure}
	\centering
	\vspace{-0.0cm}
	\setlength{\abovecaptionskip}{-2pt}
	\includegraphics[width=0.45\textwidth]{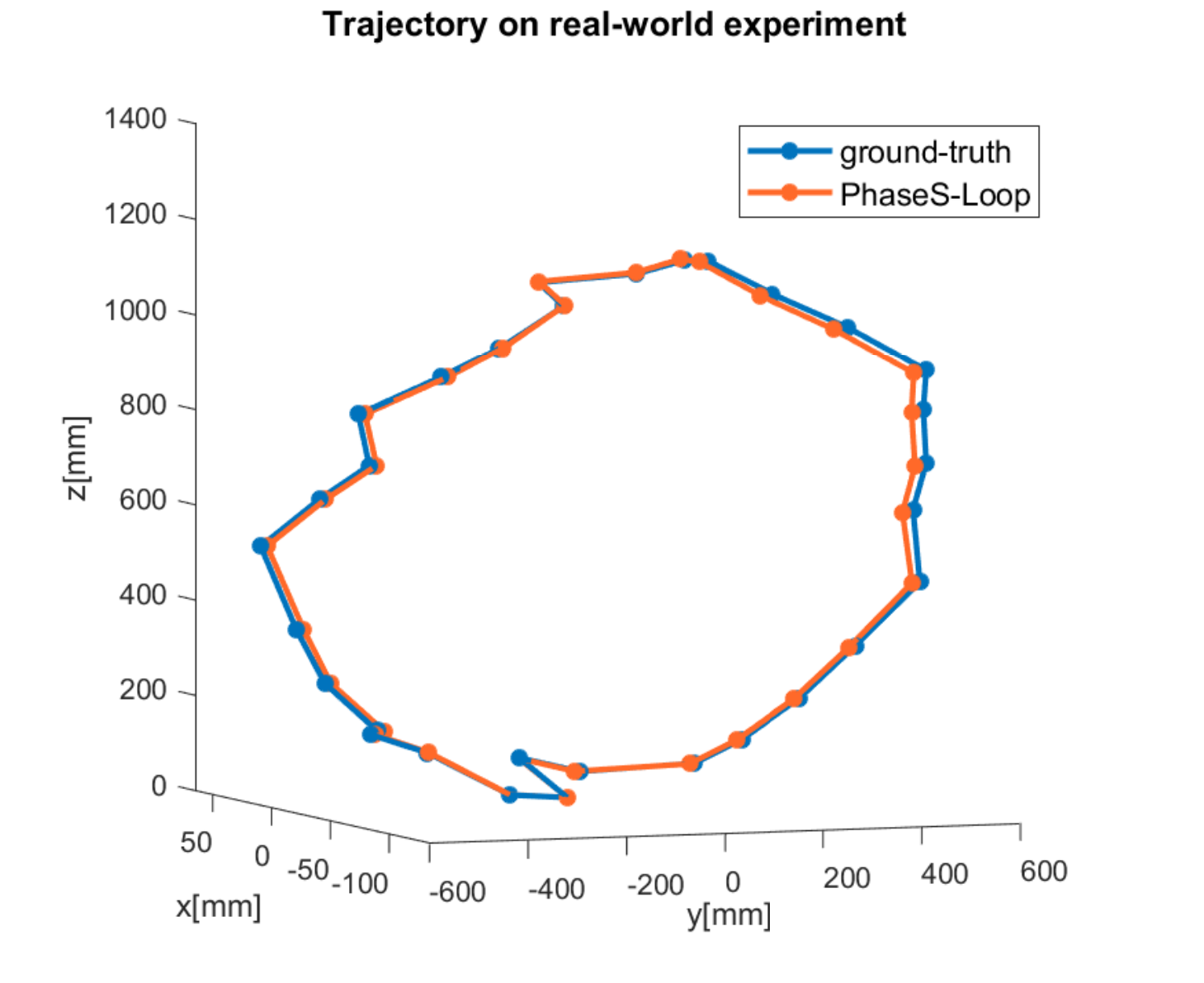}
	\caption{The plot of the estimated sensor trajectory by using the full pipeline of the proposed Phase-SLAM on David-6DoF. The ground-truth is obtained via the UR5 robotic arm.}
	\label{fig:Trajectory}
\end{figure}

\begin{figure}
	\centering
	\vspace{-0.0cm}
	\setlength{\abovecaptionskip}{-2pt}
	\includegraphics[width=0.42\textwidth]{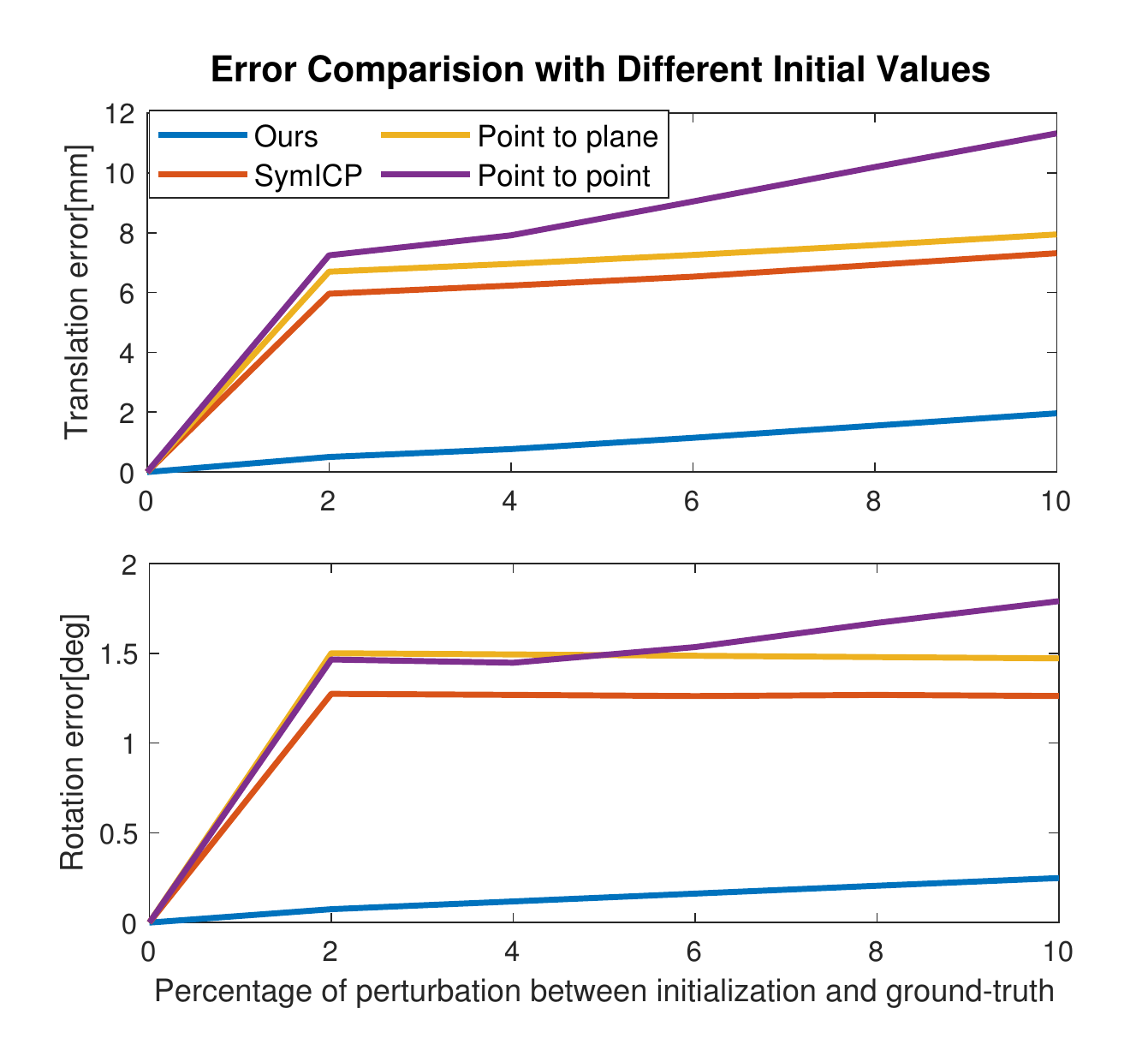}
	\caption{The plot of registration errors of (top) translation and (bottom) rotation with respect to different sensor pose initializations. The x-axis is the percentage of perturbation for pose initialization with respect to the ground-truth.}
	\label{fig:Initial}
\end{figure}

\begin{figure}
	\centering
	\vspace{-0.0cm}
	\setlength{\abovecaptionskip}{-2pt}
	\includegraphics[width=0.42\textwidth]{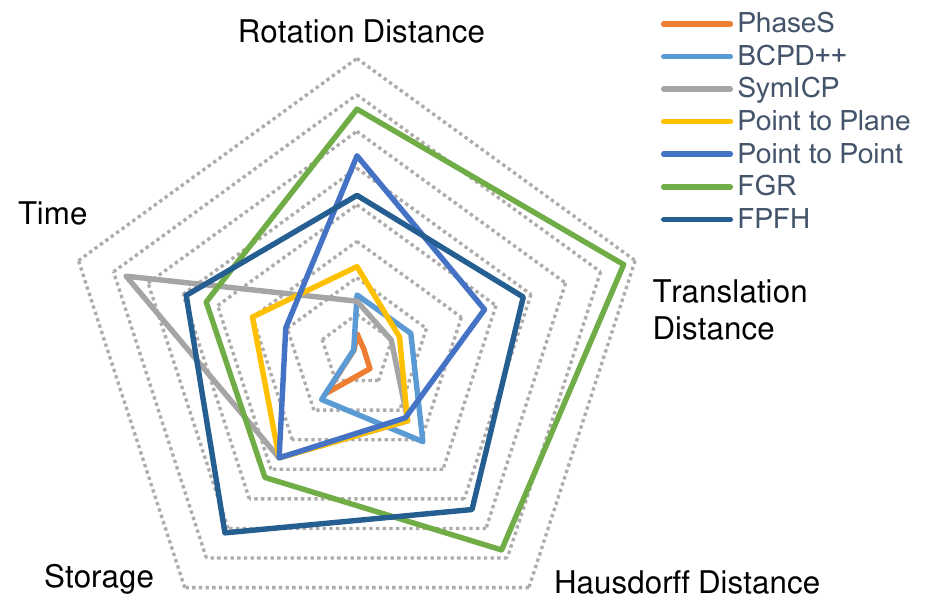}
	\caption{The radar chart of 5 performance metrics for 7 different algorithms. The rotation and translation errors are measured via the Euler distances; the Hausdorff distance is used to measure the dissimilarity between two point clouds.}
	\label{fig:star}
\end{figure}

Fig. \ref{fig: RPEReal} shows the RPE results of five different methods using the David-6DoF dataset. It is clear that the proposed method (PhaseS-Loop) has a better performance than other four methods. 
Table \ref{tab:RMSEreal} is the RMSE of ATE and the computation time for four different datasets using eight different methods.
It can be seen that the proposed method outperforms other six methods in both terms of accuracy and computation time.
The 3D object reconstruction results using the proposed method under real-world datasets are shown in Fig. \ref{fig: PCall} (d-f).
\begin{table}[h]
	\centering
	\vspace{-0.0cm}
	\caption{RMSE of ATE (mm) / Computation Time (s)}
	\renewcommand\tabcolsep{2.0pt}
	\begin{tabular}{c c c c c} 
		\toprule[1.5pt]
		Method         & David-6DoF   & David-3DoF     & PiKaChu    & Sona\\
		\midrule[1pt]
		PhaseS-Loop    & \textbf{4.69/4.20} & \textbf{4.71/3.19} & \textbf{2.09/3.30} & \textbf{1.83/3.18} \\
		PhaseS         & \textbf{6.12/4.17} & \textbf{5.74/3.17} & \textbf{3.27/3.27} & \textbf{2.29/3.15} \\
		BCPD++         & 244.41/2.79       & 53.01/2.93         & 21.72/3.72         & 22.39/3.81\\
		SymICP         & 99.28/374.26         & 28.78/345.25          & 35.68/268.69         & 30.88/242.12\\
		Point to Plane & 101.5/168.53         & 33.66/152.13         & 33.97/119.23          & 36.65/109.15\\
		Point to Point & 170.2/118.42         & 89.28/101.54        & 70.36/81.34          & 84.44/77.9\\
		FGR            & 282.3/238.25        & 224.21/213.15        & 231.17/302.45        & 149.92/191.7\\
		FPFH           & 109.85/386.8        & 95.51/153.66         & 254.34/202.14        & 90.81/217.9\\  
		\bottomrule[1.5pt]
	\end{tabular} 
	\begin{tablenotes}
	\item[1] RMSE of ATE: The root mean squared error of absolute trajectory error.
    \end{tablenotes}
	\label{tab:RMSEreal}
\end{table}
Fig. \ref{fig:Trajectory} illustrates the estimated SLI sensor trajectory and the ground truth under David-6DoF dataset, where the total trajectory length is 3.967m.
Fig. \ref{fig:Initial} shows the pose estimation errors by using local methods (SymICP, Point-to-Plant and Point-to-Point ICP) and our method without global optimizaiton under different initial values.
We can see that the proposed method is least sensitive to initial values.
Fig. \ref{fig:star} shows a radar chart of seven methods without global optimization for all datasets using 
five performance metrics (Hausdorff distance, computation time, translation/rotation errors, and storage space) for a comprehensive evaluation.
The Hausdorff distance is used to describe the dissimilarity between reconstructed point clouds and the ground-truth \cite{taha2015efficient}. 
It is obvious that the proposed method has the superior performance in all those metrics.
 \vspace{-0.0cm}
 \vspace{-0.0cm}

\section{Conclusion}

This paper presents a phase based Simultaneous Localization and Mapping (Phase-SLAM) pipeline 
for fast and accurate SLI sensor pose estimation and 3D object reconstruction.
The proposed reprojection model and local pose optimizer can achieve the odometry functionality with high efficiency, accuracy and low sensitivity to initial pose knowledge.
The proposed compressive loop closure detection technique can reduce both the loop closure computational time and data storage space.
Even without global optimization, the proposed local data registration method outperforms six other existing 3D point cloud based methods in terms of sensor pose estimation accuracy, storage space, computation time and 3D reconstruction errors.
The code of our framework and the dataset in use are available online.

\section*{Appendix}
The analytic expression of the Jacobian of $\mathbf{e}$ with respect to $\delta x, \delta y, \delta z, \delta \alpha, \delta \beta, \delta \gamma$ is provided in this section. The intermediate terms are given by
	\begin{align}
	\begin{split}
	g_x &= \displaystyle{{\partial \Phi_{k+1}}/{\partial \mu_{k+1}}}, 
	g_y = \displaystyle{{\partial \Phi_{k+1}}/{\partial \nu_{k+1}}}\\
	K &= H_p / (2\pi), s_{k+1} = m_{31} x' + m_{32} y' + m_{33} z'\\
	\mu_1 &= m_{11} - m_{31}\mu^c_{k+1}, \nu_1 = m_{21} - m_{31}\nu^c_{k+1} \\
	\mu_2 &= m_{12} - m_{32}\mu^c_{k+1}, \nu_2 = m_{22} - m_{32}\nu^c_{k+1} \\
	\mu_3 &= m_{13} - m_{33}\mu^c_{k+1}, \nu_3 = m_{23} - m_{33}\nu^c_{k+1} \\
	J_{x\alpha} &= R_{13} y - R{12} z, J_{y\alpha} = R_{23} y - R_{22} z \\
	J_{z\alpha} &= R_{33} y - R_{32} z \\
	J_{\mu\alpha} &= {\mu_1 J_{x\alpha} + \mu_2 J_{y\alpha} + \mu_3 J_{z\alpha}}/{s_{k+1}}\\
	J_{\nu\alpha} &= {\nu_1 J_{x\alpha} + \nu_2 J_{y\alpha} + \nu_3 J_{z\alpha}}/{s_{k+1}}\\
	J_{x\beta} &= -x sin\delta\beta cos\delta\gamma - y sin\delta\alpha cos\delta\beta cos\delta\gamma \\
	&- z cos\delta\alpha cos\delta\beta cos\delta\gamma \\
	J_{y\beta} &= -x sin\delta\beta sin\delta\gamma - y sin\delta\alpha cos\delta\beta sin\delta\gamma \\
	&- z cos\delta\alpha cos\delta\beta sin\delta\gamma \\
	J_{z\beta} &= -x cos\delta\beta - y sin\delta\alpha sin\delta\beta  
	- z cos\delta\alpha sin\delta\beta \\
	J_{\mu\beta} &= \mu_1 J_{x\beta} + \mu_2 J_{y\beta} + \mu_3 J_{z\beta}/{s_{k+1}}\\
	J_{\nu\beta} &= \nu_1 J_{x\beta} + \nu_2 J_{y\beta} + \nu_3 J_{z\beta}/{s_{k+1}}\\
	J_{x\gamma} &= \delta y - y', J_{y\gamma} = x'- \delta x, J_{y\gamma} = 0\\
	J_{\mu\gamma} &= \mu_1 J_{x\gamma} + \mu_2 J_{y\gamma} + \mu_3 J_{z\gamma}/{s_{k+1}}\\
	J_{\nu\gamma} &= \nu_1 J_{x\gamma} + \nu_2 J_{y\gamma} + \nu_3 J_{z\gamma}/{s_{k+1}}\\
	\end{split}
	\end{align}
	The analytic expression of Jacobian is then given by
	\begin{align} \label{sub_J}
	\begin{split}
	{\partial \mathbf{e}}/{\partial{\delta x}} &= -(g_x {\mu_1} + 
	g_y{\nu_1})/{s_{k+1}} \\
	{\partial \mathbf{e}}/{\partial{\delta y}} &=
	{f_p}/(Kz') - (g_x {\mu_2} + 
	g_y{\nu_2})/{s_{k+1}}  \\
	{\partial \mathbf{e}}/{\partial{\delta z}} &=
	{f_py'}/(Kz'^2) - (g_x {\mu_3} + 
	g_y{\nu_3})/{s_{k+1}} \\
	{\partial \mathbf{e}}/{\partial{\delta \alpha}} &=
	{f_p (J_{y\alpha}z' - J_{z\alpha}y')}/(Kz'^2) 
	- (g_x J_{\mu\alpha} + g_y J_{\nu\alpha}) \\
	{\partial \mathbf{e}}/{\partial{\delta \beta}} &=
	f_p (J_{y\beta}z' - J_{z\beta}y')/(Kz'^2) 
	- (g_x J_{\mu\beta} + g_y J_{\nu\beta}) \\
	{\partial \mathbf{e}}/{\partial{\delta \gamma}} &=
	f_p (J_{y\gamma}z' - J_{z\gamma}y')/(Kz'^2) 
	- (g_x J_{\mu\gamma} + g_y J_{\nu\gamma}) \\
	\end{split}
	\end{align}

	\addtolength{\textheight}{-12cm}   
	
	\bibliographystyle{IEEEtran}
	\bibliography{egbib}
	
\end{document}